\documentclass[11pt,a4paper]{article}
\usepackage{graphicx}
\usepackage{xcolor}
\graphicspath{ {./images/} }
\usepackage{times,latexsym}
\usepackage{url}
\usepackage[T1]{fontenc}
\usepackage{tabularx}
\usepackage{multirow}
\usepackage{adjustbox}
\usepackage[final]{acl2023}
\usepackage{mathtools}

\usepackage{natbib}
\usepackage{xspace,mfirstuc,tabulary}
\title{Semantics or spelling? Probing contextual word embeddings with orthographic noise}

\author{%
    Jacob A. Matthews \hspace{5pt} John R. Starr \hspace{5pt} Marten van Schijndel \\
    Cornell University \\
    \texttt{\{jam963, jrs673, mv443\}@cornell.edu} \\
}

\date{}

\begin{document}
\maketitle
\begin{abstract}
Pretrained language model (PLM) hidden states are frequently employed as contextual word embeddings (CWE): high-dimensional representations 
that encode semantic information given linguistic context. Across many areas of computational linguistics research, similarity between CWEs is interpreted as semantic similarity. However, it remains unclear exactly what information is encoded in PLM hidden states. We investigate this practice by probing PLM representations using minimal orthographic noise. We expect that if CWEs primarily encode semantic information, a single character swap in the input word will not drastically affect the resulting representation, given sufficient linguistic context. 
Surprisingly, we find that CWEs generated by popular PLMs are highly sensitive to noise in input data, and that this sensitivity is related to subword tokenization: the fewer tokens used to represent a word at input, the more sensitive its corresponding CWE. This suggests that CWEs capture information unrelated to word-level meaning and can be manipulated through trivial modifications of input data. We conclude that these PLM-derived CWEs may not be reliable semantic proxies, and that caution is warranted when interpreting representational similarity. 

\end{abstract}

\begin{table*}[t]
	\centering
	\begin{tabular} {
			| m{10em} | m{4em} | m{4em} | m{7em} | m{13em} | }
		\hline
		Model                      & Word       & Edited & Word Tokens            & Edited Tokens                       \\
		\hline
		GPT-2       &            &                  & ``contenders''       & ``cont'', ``e'', ``ld'', ``ers'' \\
		BERT  & contenders & contelders            & ``contender'', ``s'' & ``con'', ``tel'', ``ders''       \\
		XLNet &            &             & ``contenders''       & ``con'', ``tel'', ``der'', ``s'' \\
		\hline
	\end{tabular}
	\caption{Example of the effect of noise on model tokenization. The second ``n'' in ``contenders'' has been replaced with an ``l''. Though GPT-2 and XLNet tokenize the original word identically, a single character replacement results in distinct splitting behavior in the edited tokenization.} 
	\label{tab:noise_example}
\end{table*}

\section{Introduction}
Contextual word embeddings (CWE) have become commonplace, enabling a wide range of
research and industry applications. These representations, typically generated using transformer language 
models like BERT \citep{devlin2019bert}, differ from their non-contextual predecessors like word2vec \citep{Mikolov2013EfficientEO} in that these embeddings vary based on the linguistic context of the words they represent. However, the extent to which these representations capture contextual \citep{ethayarajh-2019-contextual}, linguistic \citep{Miaschi2020ContextualAN}, and social \citep{Kurita2019QuantifyingSB, Guo2020DetectingEI} information remains an area of significant interest. 
In emerging subfields like semantic shift detection (SSD), CWEs are frequently employed as semantic proxies \citep{Montanelli2023ASO}. BERT or similar pretrained language models (PLMs) are commonly used to generate CWEs which function as continuous representations of semantic information. This use of CWEs assumes that PLMs encode semantic information in their hidden states, and that the similarity between CWEs can be directly interpreted as a measure of semantic similarity. 

We investigate this assumption using a simple probing task.\footnote{All code and data are available at \url{https://github.com/jam963/semantics-or-spelling/}.} We generate CWEs of alphabetic words in authentic contexts, then we compare them pairwise to CWEs of the same word with a single character swapped for another of the same case. Since the rest of each word and context sequence is otherwise kept identical, we would expect that the unedited characters and linguistic context should be sufficient to generate extremely similar CWEs to their unedited counterparts. Put differently, if CWEs primarily capture semantic information, they should be robust to minor orthographic noise.

In fact, this is not the case. We observe that CWEs are quite sensitive to minor orthographic noise, regardless of the PLM used to generate them. Moreover, this effect is strongest for single token words, which account for over a third of English words in our test data. We find that these representations are highly dependent on tokenization: words represented by fewer tokens are more sensitive to orthographic noise. Changes at the character level result in drastic differences in token-level representation, which in turn causes the model to produce distinct CWEs.  

\section{Background} 
While some prior work has addressed the robustness of PLMs to noisy 
input sequences \cite{xue-etal-2022-byt5, niu-etal-2020-evaluating, karpukhin-etal-2019-training}, 
these studies rely on downstream task performance to quantify noise robustness. However, to our knowledge, there is no
thorough account of \emph{representational} noise robustness in PLMs. Additionally, there are known problems with representations generated by PLMs, such as anisotropy \cite{mimno-thompson-2017-strange,ethayarajh-2019-contextual}, the influence of 
rogue dimensions \cite{timkey-van-schijndel-2021-bark}, and similarity underestimation due to 
frequency \cite{zhou-etal-2022-problems}.

One notable difference between non-contextual word-based embedding models such as word2vec \citep{Mikolov2013EfficientEO} or GloVe \citep{Pennington2014GloVeGV} and PLMs is tokenization. Whereas the former utilize a fixed model vocabulary, contemporary transformer-based PLMs generally rely on \textit{subword tokenization} strategies for efficiency and flexibility. Most models use a small number of very similar methods. Byte Pair Encoding 
\citep[BPE;][]{sennrich-etal-2016-neural}, specifically byte-level BPE, is used in a wide 
range of popular PLMs, including GPT-2 \citep{radford2019language}, GPT-3 \citep{brown2020language},
RoBERTa \citep{DBLP:journals/corr/abs-1907-11692}, and BLOOM \citep{Scao2022BLOOMA1}, among others. 
BPE employs a set of merge rules to convert a fixed vocabulary into a segmented 
subword vocabulary. This approach allows out-of-vocabulary words to be 
tokenized into sequences of in-vocabulary subword tokens. Besides BPE, 
there are a number of other subword tokenization methods, 
including SentencePiece \citep{kudo-richardson-2018-sentencepiece} 
and WordPiece \citep{6289079}. 

In addition, character-based \citep{tay2022charformer} and tokenizer-free methods \citep{clark-etal-2022-canine} have been investigated with transformer models, as well as regularization approaches to address known shortcomings with existing tokenizers \citep{provilkov-etal-2020-bpe}. For a comprehensive overview and taxonomy of contemporary tokenizers, see \citet{mielke2021between}.
 
\section{Data and model selection}
For our analyses, we establish a test vocabulary by isolating every unique, alphabetic, whitespace-delimited 
word of more than 3 characters in the \texttt{Wikitext-2-raw-v1} training set \citep{merity2016pointer}. This results in a 
total vocabulary size of 68,725 alphabetic words. To generate edited words, we randomly select a character in each word and replace it with another random alphabetic character of the same case (Table~\ref{tab:noise_example}). We will refer to these as \textit{edited words}.

The PLMs we analyze represent a range of 
families, sizes, and types, all of which are freely 
available via the HuggingFace \texttt{transformers} package \citep{wolf2020huggingfaces}, which we also 
use here for our analyses. We generate CWEs using versions of GPT-2 \citep{radford2019language}, BERT \citep{devlin2019bert}, RoBERTa \citep{Liu2019RoBERTaAR}, XLNet \citep{Yang2019XLNetGA} and BLOOM \citep{Scao2022BLOOMA1}. While GPT-2 and BLOOM are causal language models which are typically not used for generating CWEs, they represent a class of extremely popular and widely-available PLMs. Additional model details are described in the Appendix (Table~\ref{plms}).

\begin{figure*}[t]
	\centering
	\includegraphics[width=\textwidth]{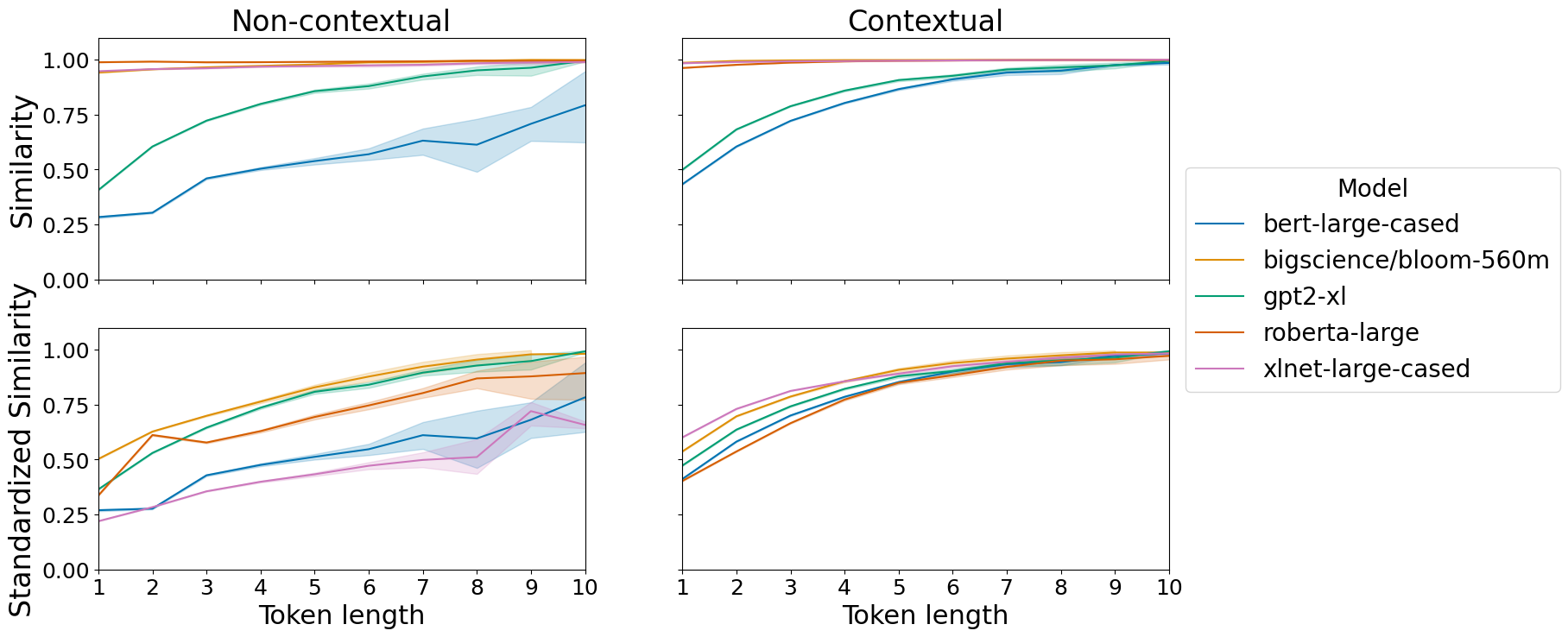}
	\caption{Similarities between CWEs and edited-CWEs, using cosine (top) and Spearman (bottom) for individual words without contexts (left) and with contexts (right). The shaded area around each line represents the 99\% confidence interval. }
	\label{fig:similarities_all}
\end{figure*}

\section{Measuring similarity of CWEs} \label{rep1}
We use two measures of representational similarity between CWEs: cosine similarity and Spearman correlation. 
We first define a mean-pooled word embedding $\bar{\mathbf{w}}$ as 
\begin{equation}
\bar{\mathbf{w}}=\frac{1}{n} \sum_{i=1}^{n} \mathbf{h_{i}}
\end{equation}
where $\mathbf{h_{i}}$ is a hidden state corresponding to token $i$ in a word of $n$ tokens. The 
cosine similarity $cos$ between a word embedding $\bar{\mathbf{w}}$ and its noised counterpart $\bar{\mathbf{w}}_{edit}$
is given by 
\begin{equation}
cos(\bar{{\mathbf{w}}}, \bar{\mathbf{w}}_{edit}) = \frac{\bar{\mathbf{w}} \cdot \bar{\mathbf{w}}_{edit}}{\lVert\bar{\mathbf{w}}\rVert \lVert\bar{\mathbf{w}}_{edit}\rVert}
\end{equation}  
where values closer to 1.0 are interpreted as more similar and therefore more noise robust.  

Though commonly used, cosine similarity is \textit{unstandardized}. As such, small numbers of outlier or "rogue" dimensions often dominate these similarity scores \citep{timkey-van-schijndel-2021-bark, kovaleva-etal-2021-bert}. Moreover, PLMs have been shown to exhibit \textit{anisotropy}, where all representations occupy a narrow region of the embedding space \citep{mimno-thompson-2017-strange,ethayarajh-2019-contextual}, which distorts similarity measures like cosine. Thus we also calculate the Spearman rank 
correlation coefficient, $\rho$, for all embedding pairs, 
\begin{equation}
    \rho_{\bar{\mathbf{w}}, \bar{\mathbf{w}}_{edit}} = \frac{cov(\mathbf{R}(\bar{\mathbf{w}}), \mathbf{R}(\bar{\mathbf{w}}_{edit}))}{\sigma_{\mathbf{R}(\bar{\mathbf{w}})}\sigma_{\mathbf{R}(\bar{\mathbf{w}}_{edit})}} 
\end{equation}
where $cov(\mathbf{R}(\bar{\mathbf{w}}), \mathbf{R}(\bar{\mathbf{w}}_{edit}))$ is the covariance of the rank-transformed CWEs $\bar{\mathbf{w}}$ and $\bar{\mathbf{w}}_{edit}$. 
Because the rank transformation mitigates the effects of rogue dimensions and anisotropy, Spearman is a more robust %
similarity measure than cosine \citep{zhelezniak-etal-2019-correlation}. %

\begin{figure*}[t]
        \centering
	\includegraphics[width=\textwidth]{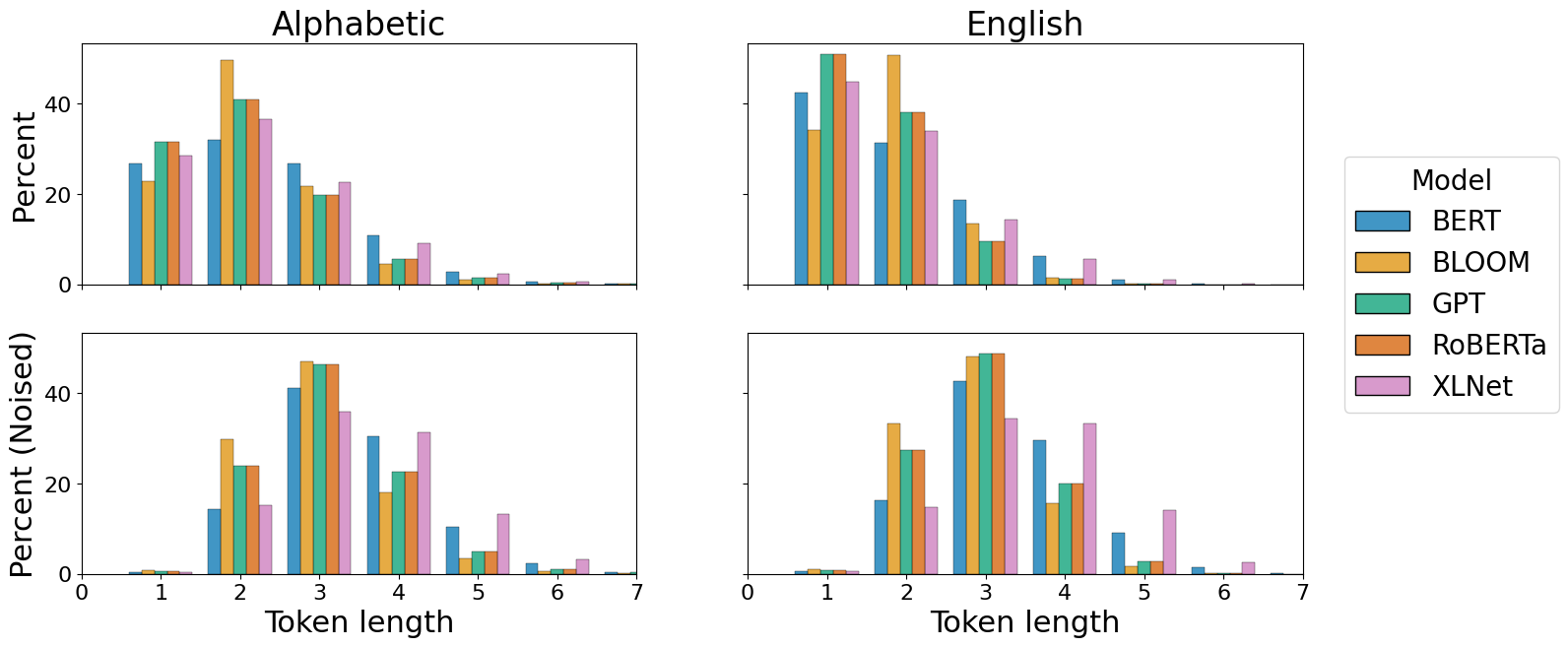}
	\caption{Distribution of token lengths for all alphabetic words (left) and English words (right), tokenized 
         without (top) and with noise (bottom). For most models, English words tend to be tokenized with fewer tokens (top right). Noise tends to increase the token length, with most noisy words requiring 3 or more tokens instead of 1 or 2.}
	\label{fig:tok_len_dist}
\end{figure*}

\section{Representational similarity analyses}
\subsection{Non-contextual similarity}\label{non-contextual}
To determine the baseline influence of our editing procedure on word embeddings without linguistic context, we first generate
non-contextual word embeddings for every word and edited word. We pass each word and corresponding
edited word to all models independently, without any additional context, then mean pool
the final layer hidden states corresponding to the tokens in each word. This gives us a pair of word embeddings for all
vocabulary items $(\mathbf{w}, \mathbf{w}_{edit})$, which we compare using cosine similarity and Spearman correlation. For both similarity measures, values closer to 1.0 suggest that a model has similar representations for both the edited and unedited words.

\subsubsection{Results}
We report results from the largest models tested in Figure~\ref{fig:similarities_all} (Left), with similarity scores binned by the number of tokens used to represent the unedited word, or the \textit{token length}. While some models (BLOOM, GPT2-small, and the RoBERTa models) have average cosine similarities close to 1.0 for all token lengths, standardized similarities (Spearman) are universally low for all but the highest number of tokens, falling between 0.2 and 0.5 for single token words.\footnote{See Figures \ref{fig:appendix_standardized} and \ref{fig:appendix_cosine} in Section \ref{appendix:results} for all model results.}

\subsection{Contextual similarity} 
Since CWEs are designed to modulate their representations according to linguistic context, we may expect that context would correct these representational discrepancies, allowing the model to better ignore minor orthographic noise. 

To simulate the conditions under which these models are typically employed, we investigate how noise affects CWE representations given authentic linguistic contexts. We repeat our analysis from Section~\ref{non-contextual}, but provide contexts from Wikitext
for each word when generating hidden layer representations. We find naturally occurring 100-word contexts that contain each unedited test word, and we pool hidden layer representations in that context to generate CWEs corresponding to the test word and their edited twin. Thus, between a given word and edited word, their respective 100-word input sequence varies only by a single character. As in section \ref{non-contextual}, we compare the resulting CWEs for the original and edited items using cosine similarity and Spearman correlation.

\subsubsection{Results}\label{context-results}
Though context notably increases similarity and standardized similarity for some models like BERT and XLNet, it has almost no effect on BLOOM, RoBERTa, and larger GPT-2 variants (Figure~\ref{fig:similarities_all}, Right; see also Figures \ref{fig:appendix_standardized} and \ref{fig:appendix_cosine} in Appendix). Just as in the non-contextual case, words originally represented by fewer tokens are the least noise robust (Spearman averaging 0.5 for single token words). The fact that similarity scores approach 1.0 at high numbers of tokens indicates that PLMs are in principle able to use context to mitigate input noise; however, the dependence of subword tokenization schemes on occurrence frequencies seems to inhibit this behavior and make models more susceptible to orthographic noise. %

\section{PLM token length distributions}
To study the impact of frequency on the behaviors we've found, we analyze the distribution of word lengths (in tokens) in our test vocabulary (Figure~\ref{fig:tok_len_dist}, Top). 
Because the Wikitext dataset contains non-English words, we also use the MorphoLex database \citep{SnchezGutirrez2018MorphoLexAD} to identify
authentic English words (40,013 words).
Though specific tokenizer implementations vary somewhat between models, we find that the token 
length distributions are generally similar (Figure \ref{fig:tok_len_dist}): as many as 90\% of alphabetic words are less than 3 tokens in length, 
while English words are mostly composed of even fewer tokens, usually only one or two. Coupled with our results from section \ref{context-results}, 
this suggests that most English words will exhibit very low robustness to minor orthographic noise, even when observed in long linguistic contexts.

Based on how subword tokenizers work, edited words tend to be represented by more tokens than unedited words (Figure~\ref{fig:tok_len_dist}, Bottom). Though subword tokenizers can accommodate orthographic noise, they can only do so by combining multiple tokens that are not present in the unedited word. For all but very rare words composed of many tokens, the mismatch in token-level representation is too much for context to representationally ``repair,'' even when the noise in question only consists of a single character. %

\section{Discussion}
We show that contextual word embeddings (CWEs) are highly sensitive to orthographic noise and that this effect is related to the way text data is processed by subword tokenizers. Our results indicate that %
CWE representations can be manipulated through trivial orthographic changes to the input -- given 100 words of context, 40\%-60\% of a word's semantic identity is lost when a single character is changed -- challenging the assumption that CWEs capture primarily word-level semantic information. In addition, this study suggests that CWEs and representational similarity measures are surprisingly fragile, and common noise sources such as misspellings and optical character recognition errors may result in unpredictable model behavior downstream. As such, we advise caution when relying on PLM-generated CWEs as semantic proxies in computational research, especially when dealing with noisy data. 

\section{Limitations}
This study relies on third-party packages and open-source models which may differ from the implementation described in cited literature. Our procedure for generating CWEs, designed to be compatible with a range of models, may not be optimal, as we only combine the last hidden states through mean pooling. Though we believe the number of words in our test data to be sufficient, the vocabulary in Wikitext is still limited and may not be representative of English words in all contexts.   

We note two limitations of our noising procedure, which 1) operates on random characters and 2) applies to all classes of words. Regarding the first limitation, our procedure may not reflect the actual distribution of orthographic noise in human-generated training data, which may already have some systematic character-level noise present (i.e. ``teh'' for ``the'', swapping ``j'' for ``h''). However, we believe that our approach provides a generalizable view of PLM behavior rather than a focused study of identifiable noise sources. Because these models are deployed across a vast range of linguistic contexts, we do not take a strong view of what constitutes ``authentic'' orthographic noise.

With respect to the second limitation, our study does not address the influence of part-of-speech or other word-level characteristics on CWE noise robustness. By limiting our investigation to words with over 3 characters, our study does not address the noise robustness of many function words in English.

\bibliography{tacl2021}
\onecolumn
\appendix
\section{Appendix}
\subsection{Model details}\label{appendix:models}
\begin{table}[h!]
	\centering
        \small
	\begin{adjustbox}{width=\textwidth}
		\begin{tabular}{|l|l|l|l|l|l|}
			\hline
			\textbf{Family}          & \textbf{Model}                     & \textbf{Parameters} & \textbf{Tokenizer} & \textbf{Training Objective} & \textbf{Released} \\ \hline
			\multirow{2}{*}{BERT}    & bert-base-cased                    & 109M                & WordPiece          & MLM, NSP                    & 2018              \\ \cline{2-3}
			                         & bert-large-cased                   & 335M                &                    &                             &                   \\ \hline
			\multirow{4}{*}{GPT}     & gpt2                               & 137M                & BPE                & CLM                         & 2019              \\ \cline{2-3}
			                         & gpt2-medium                        & 380M                &                    &                             &                   \\ \cline{2-3}
			                         & gpt2-large                         & 812M                &                    &                             &                   \\ \cline{2-3}
			                         & gpt2-xl                            & 1.61B               &                    &                             &                   \\ \hline
			\multirow{2}{*}{RoBERTa} & roberta-base                       & 125M                & BPE                & MLM                         & 2019              \\ \cline{2-3}
			                         & roberta-large                      & 335M                &                    &                             &                   \\ \hline
			\multirow{2}{*}{XLNet}  & xlnet-base-cased                   & 150M                & SentencePiece      & PermLM                         & 2019              \\ \cline{2-3}
			                         & xlnet-large-cased                  & 340M                &                    &                             &                   \\ \hline
			\multirow{1}{*}{BLOOM}   & bloom-560m                         & 560M                & BPE                & CLM                         & 2022              \\ \hline
		\end{tabular}
	\end{adjustbox}
	\caption{Summary of PLMs analyzed. Training
		objectives include masked language modeling (MLM), next-sentence
		prediction (NSP), causal language modeling (CLM), and permutation language
		modeling (PermLM).}
        \label{plms}
\end{table}

\subsection{Full representational analysis results}\label{appendix:results}

\begin{figure}[h!]
	\centering
	\includegraphics[width=\textwidth]{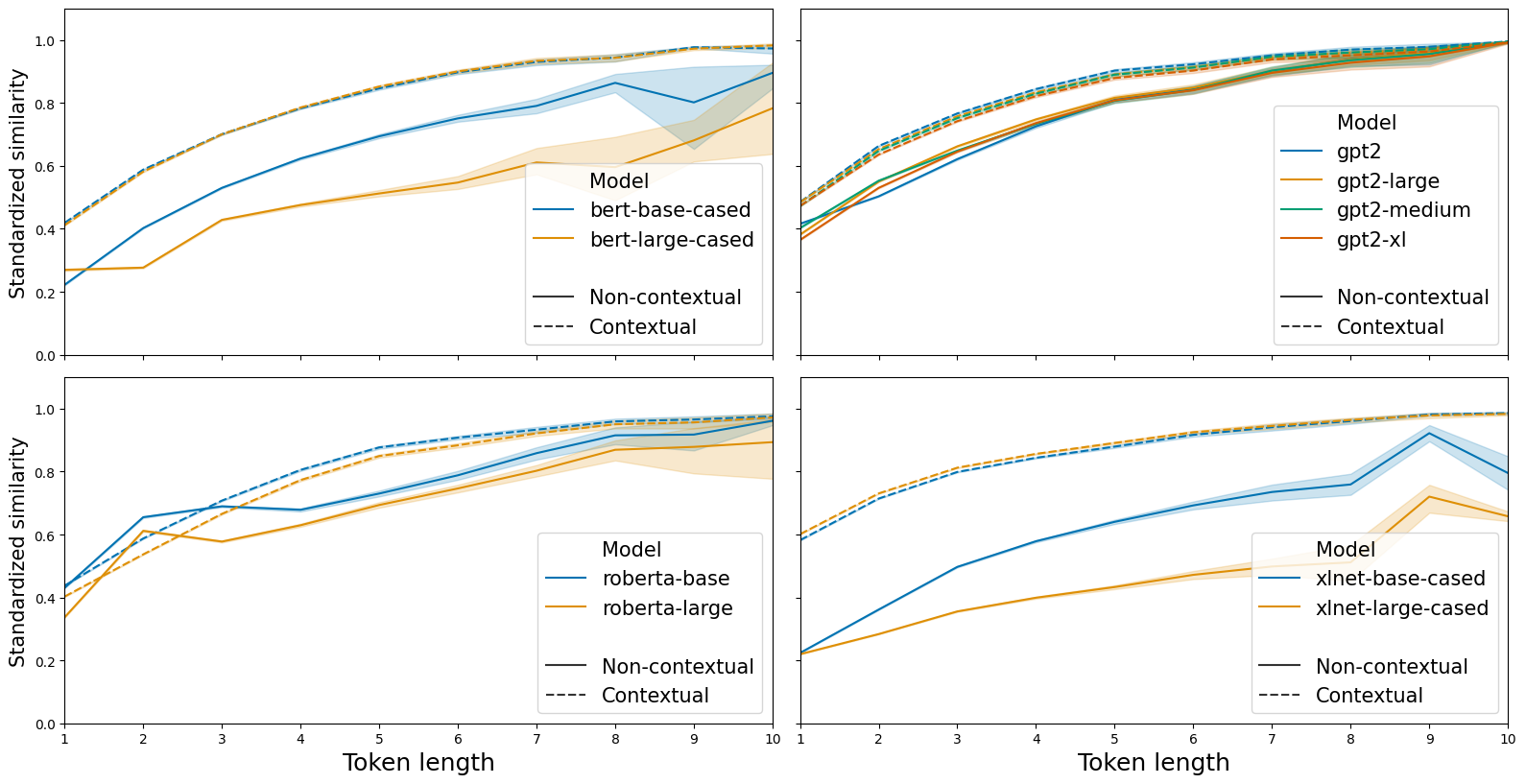}
	\caption{Standardized similarities (Spearman), grouped by model family. Token length is measured as the number of tokens used to encode the original, unedited word. We show similarities both for CWEs generated with context (dashed line) and without (solid line), and the shaded area around each line represents the 99\% confidence interval.}
	\label{fig:appendix_standardized}
\end{figure}

\begin{figure}[h]
	\centering
	\includegraphics[width=\textwidth]{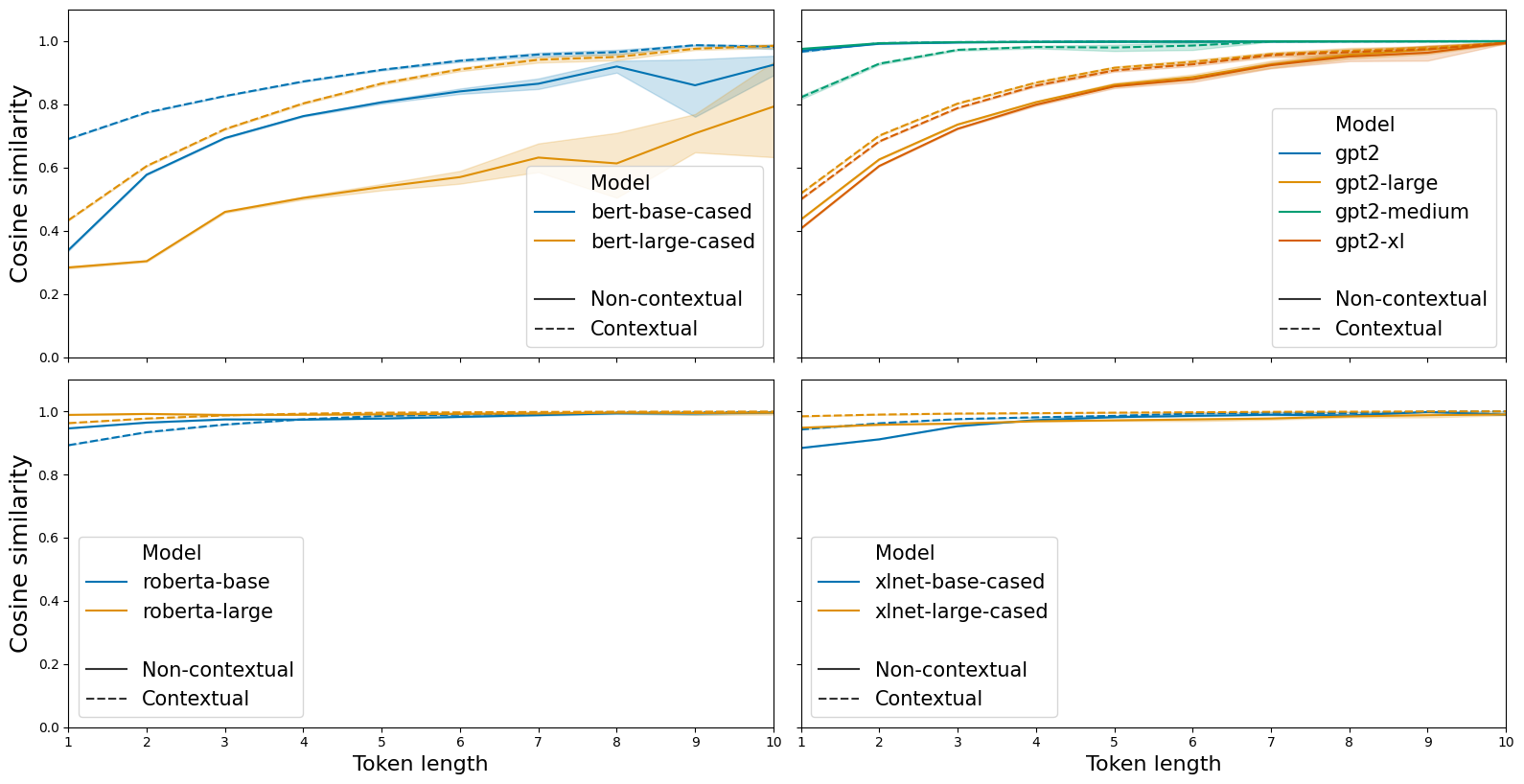}
	\caption{Cosine similarities between CWEs and edited-CWEs, grouped by model family. Formatting is otherwise identical to Figure \ref{fig:appendix_standardized}.}
	\label{fig:appendix_cosine}
\end{figure}

\end{document}